# Scalable Scientific Interest Profiling Using Large Language Models

Yilun Liang[1#], Gongbo Zhang, PhD[2#], Edward Sun[3], Betina Idnay, PhD, RN[2], Yilu Fang, MA[2], Fangyi Chen, MA[2], Casey Ta, PhD[2], Yifan Peng, PhD[4,*], Chunhua Weng, PhD[2,*]

#: equal contribution first authors;

*: equal contribution senior authors;

[1] Tandon School of Engineering, New York University, Brooklyn, NY, USA

[2] Department of Biomedical Informatics, Columbia University, New York, NY, USA

[3] Henry Samueli School of Engineering and Applied Science, University of California, Los Angeles, CA, USA

[4] Department of Population Health Sciences, Weill Cornell Medicine, New York, NY, USA

*: Correspondence: yip4002@med.cornell.edu; cw2384@cumc.columbia.edu

## Abstract

**Objective:** Research profiles highlight scientists' research focus, enabling talent discovery and fostering collaborations, but they are often outdated. Automated, scalable methods are urgently needed to keep these profiles current.

**Methods:** In this study, we design and evaluate two Large Language Models (LLMs)-based methods to generate scientific interest profiles—one summarizing researchers' PubMed abstracts and the other generating a summary using their publications' Medical Subject Headings (MeSH) terms—and compare these machine-generated profiles with researchers' self-summarized interests. We collected the titles, MeSH terms, and abstracts of PubMed publications for 595 faculty members affiliated with Columbia University Irving Medical Center (CUIMC), for 167 of whom we obtained human-written online research profiles. Subsequently, GPT-4o-mini, a state-of-the-art LLM, was prompted to summarize each researcher's interests. Both manual and automated evaluations were conducted to characterize the similarities and differences between the machine-generated and self-written research profiles.

**Results:** The similarity study showed low ROUGE-L, BLEU, and METEOR scores, reflecting little overlap between terminologies used in machine-generated and self-written profiles. BERTScore analysis revealed moderate semantic similarity between machine-generated and reference summaries (F1: 0.542 for MeSH-based, 0.555 for abstract-based), despite low lexical overlap. In validation, paraphrased summaries achieved a higher F1 of 0.851. A further comparison between the original and paraphrased manually written summaries indicates the limitations of such metrics. Kullback-Leibler (KL) Divergence of term frequency-inverse document frequency (TF-IDF) values (8.56 and 8.58 for profiles derived from MeSH terms and abstracts, respectively) suggests that machine-generated summaries employ different keywords than human-written summaries. Manual reviews further showed that 77.78% rated the overall impression of MeSH-based profiling as “good” or “excellent,” with readability receiving favorable ratings in 93.44% of cases, though granularity and factual accuracy varied. Overall, panel reviews favored 67.86% of machine-generated profiles derived from MeSH terms over those derived from abstracts.

**Conclusion:** LLMs promise to automate scientific interest profiling at scale. Profiles derived from MeSH terms have better readability than profiles derived from abstracts. Overall, machine-generated summaries differ from human-written ones in their choice of concepts, with the latter initiating more novel ideas.

## 1. Introduction

Scalable profiling of researchers' scientific interests facilitates cost-effective strategic institutional planning and decision-making [1–3]. While platforms such as Google Scholar [4], Semantic Scholar [5], ResearchGate [6], Open Researcher and Contributor ID (ORCID) [7], and the DataBase systems and Logic Programming (DBLP) [8] have become widely used to showcase academic work, most of these online researcher profiles remain outdated, inaccurate, or incomplete [9]. Notably, a recent survey [10] revealed that researchers are unsatisfied with their scientific profiles, which are often incomplete or misrepresented on ResearchGate, as they were usually constructed by scraping details from the web. Indeed, such a common approach — web scraping — for collecting researchers' information and building their profiles has limitations. The lack of current information in online scientific profiles not only misrepresents busy researchers who do not have time to manually update these profiles regularly, but also significantly hinders the identification of experts based on their most recent research focus [10]. To address this unmet need, Welke et al. [11] built an automated pipeline to profile and visualize scholars' research interests. However, it only extracts Medical Subject Headings (MeSH) terms from publication metadata and visualizes them in a word cloud without generating a narrative summary. It is neither convenient nor ideal as a surrogate for fluent, manually written research summaries, which are desired most of the time.

Recent advances in foundation models, such as BERT [12–15] and GPT [16,17], have revolutionized capabilities in text summarization [18–32]. These advancements present a novel opportunity to address the deficiencies in the methods for automatically generating profiles based on researchers' current and historical research activities. Leveraging the latest Gen AI technologies, we present a novel pipeline to enhance researcher profile creation by systematically extracting and synthesizing researchers' publications from PubMed. To ensure relevance and informativeness, we included only articles published in the past decade and on which the researcher provided significant contributions as being among the first three authors or designated as the senior author. We then employed two distinct approaches to generating researcher profiles using large language models (LLMs): 1) text summarization of publication abstracts and 2) text generation based on MeSH terms and keywords. The quality of scientific interest profiles depends on how comprehensively and accurately the profile summarizes the researcher's work and expertise while balancing specificity against abstraction [33]. A scientific interest profile that verbatim stitches original sentences from the source documents is considered lower quality compared to those with proper abstraction and summarization. Recent evidence suggests that writers with writing assist from A.I. usually have homogenized language and ideas, with their essays converge on similar n-grams, topics, and phrasings; such observation raises concerns about loss of originality in writings [34, 35]. These studies reflect the intrinsic nature of lack of originality in A.I. generated writings, which, in turn, motivates our focus on semantic richness/novelty rather than lexical overlap alone. With this consideration, we also propose a new metric, which utilizes the Kullback-Leibler (KL) divergence [36] between term frequency-inverse document frequency [37] (TF-IDF) value distributions of the compared content to quantify and characterize the differences of the vocabulary patterns between machine-generated and self-written profiles.

This study makes the following original methodological contributions. First, we presented a novel pipeline that automatically creates researcher profiles by systematically extracting data from PubMed and filtering data based on authorship position and publication recency. Then we designed and compared two LLM-based profile generation strategies. On this basis, we analyzed profile quality in terms of content similarity and semantic richness. We further proposed a KL divergence-based metric that quantifies the vocabulary distribution shift between human-written research profiles and machine-generated profiles, offering a proxy for measuring LLM's ability for text abstraction and the semantic richness of the resulting summary.

**Statement of Significance**

| | |
|---|---|
| **Problem** | The lack of up-to-date information in online scientific profiles not only misrepresents busy researchers who lack the time to manually maintain these profiles but also hinders timely and accurate identification of scientific experts based on their most recent research focus. |
| **What is Already Known** | Large language models promise to improve the accuracy and efficiency for text summarization. |
| **What this Paper Adds** | We developed a novel scalable pipeline to automatically retrieve relevant PubMed data and metadata for individual researchers. We also introduced a KL divergence-based metric to qualify and quantify the differences in the selection of concepts between human-written research profiles and machine-generated profiles. |
| **Who Would Benefit** | Researchers interested in scale scientific profiling using large language models. Academic institutions and research offices seeking up-to-date expert directories; funding agencies and collaborators seeking to identify experts; and bibliometric service providers looking to scale scientific profile generation. |

## 2. Methods

We created a pipeline to acquire human-written research summaries from the Web and automatically summarize researchers' scientific profiles. It consists of three components (Figure 1): (1) data collection, (2) model development, and (3) evaluation and analysis.

### 2.1 Data Collection

For methodology illustration, we collected data on all faculty members in the Columbia University Vagelos College of Physicians and Surgeons because their websites are well-organized, feature a uniform HTML structure, and contain self-summarized research interests. We then used BeautifulSoup [38] and Selenium [39] to extract each researcher's name, affiliation, and research interest overview from their official web pages. Finally, we used the National Institutes of Health (NIH) Entrez Programming Utilities (E-utilities) [40] to download the titles, abstracts, and MeSH terms of the researchers' publications from PubMed. Moreover, for summarization, we only included the publications where the scholars were among the first three or last three authors (Table 1), prioritizing the work contributed primarily by the researchers. For scholars with common names, we have added the institutional affiliation to facilitate name disambiguation. We excluded faculty members with empty self-summarized research interests or no published articles. A total of 595 faculty members were included in the data collection phase.

Table 1: Comprehensive Table Summary of the Background Statistics of Collected Researchers

| | |
|---|---|
| **Number of researchers, n** | 167 |
| **Gender** (F/M) | |
| Female | 116 |
| Male | 52 |
| **Academic rank** | |
| Professor | 105 |
| Associate Professor | 28 |
| Assistant Professor | 34 |
| **Areas** | |
| Biochemistry and Molecular Biophysics | 36 |
| Neuroscience | 25 |
| Genetics and Development | 24 |
| Microbiology and Immunology | 23 |
| System Biology | 17 |
| Molecular Pharmacology and Therapeutics | 13 |
| Biomedical Informatics | 11 |
| Physiology and Cellular Biophysics | 8 |
| Medical Humanities and Ethics | 4 |
| Biostatics | 2 |
| Medicine | 2 |
| School of Nursing | 1 |
| Psychiatry | 1 |
| **Profiles word count** | |
| 0-99 | 23 |
| 100-199 | 60 |
| 200-299 | 41 |
| 300-399 | 17 |
| >=400 | 26 |
| **Number of publications** | |
| 0-29 | 41 |
| 30-59 | 56 |
| 60-89 | 29 |
| 90-119 | 12 |
| >=120 | 29 |

## 2.2 Model Development

We explored two strategies for generating researcher profiles using LLMs. The first strategy inputted publication keywords into the model without providing additional context or data processing (MeSH-based). We categorized the keywords into two groups: methodology and health domains. We requested the LLMs to summarize each domain separately. The second method used the “Divide-and-Conquer” [41] approach, where the model was fed with publication abstracts to summarize the context (abstract-based). GPT-4o-mini [42] enforces a limit of 128,000 tokens for the input, which is insufficient to fit the content of all abstracts for senior scholars with hundreds of publications. To overcome this challenge, we first applied Latent Dirichlet Allocation (LDA) [43] to group the publication records by topic. Then, publications under each topic were condensed into succinct paragraphs, which were later combined for a final round of summarization. To ensure that the GPT-4o-mini model consistently generated researcher profiles like human-written ones, the model was provided with a single example, which included instructions for profiling, MeSH terms or abstracts, and the human-written profiles for the corresponding research summary (Figure 2). Figure 3 shows example profiles for one researcher, including a) MeSH-based and b) abstract-based profiles, c) paraphrased human-written profiles, and d) human-written profiles. For these tasks, we used GPT-4o-mini as the backbone. We generated researcher profiles for the 595 researchers collected. We primarily selected GPT-4o-mini because it was the state-of-the-art Large Language Model available at the time of our research, offering advanced text summarization and generation capabilities. Also, its affordability and speed allow us to efficiently generate many summaries and facilitate a scalable evaluation of our pipeline. Therefore, the balance between advanced performance and affordability made GPT-4o-mini suitable for the task of systematically generating research summaries at scale.

### 2.3 Machine Evaluation

Human-written research summaries, or human-written profiles, are required for machine evaluation as automatic metrics need human-written profiles for comparison to generate meaningful results. Therefore, for the machine evaluation phase, we applied a filter to the 595 researchers collected, excluding those researchers with an empty human-written research summary. A total of 167 researchers, with human-written research summaries, were included in the machine evaluation phase.

#### 2.3.1 Natural Language Generation (NLG) Metrics

We performed both quantitative and qualitative analyses for the LLM-generated researcher profiles. The lexical metrics include ROUGE-L [44], BLEU [45], and METEOR [46], which are widely used to measure the similarity of word choices between source and target texts. BLEU focuses on lexical precision; ROUGE emphasizes lexical recall; and METEOR balances precision and recall, while incorporating synonyms and word order for a more nuanced lexical assessment. In addition, we used LLMs to paraphrase the human-written research profiles, which served as the baseline for assessing the effectiveness of the evaluation metrics. Specifically, when MeSH-based and abstract-based approaches were evaluated using these metrics, their generated profiles were compared against paraphrased ones. We conducted paired t-tests ($\alpha = 0.05$) on the score differences between system outputs.

#### 2.3.2 Semantic Richness Metrics

Prior studies have shown that traditional NLG metrics often fail to capture the semantics of the text content [47–49]. The semantics are typically reflected in the keywords of documents, which can be reflected in term frequency. Based on this intuition, we introduced a new metric based on TF-IDF to assess the uniqueness of word choices relative to the overall corpus, and KL divergence, which measures the difference between two distributions. Taking the KL divergence of the TF-IDF quantifies the vocabulary distributional differences between the two documents. We incorporated these measures between the machine-generated and human-written profile texts to assess the semantic richness, as the ability to coin new content or terms in research profiles. Motivated by the report that AI-assisted writing show homogeneity in n-grams and topics, we interpret lower KL divergence and fewer TF-IDF unique terms as evidence of reduced novelty and of greater homogenization [34]. To focus on informative words, we eliminated stop words—commonly used words carrying little information like "the" and "and"—from the texts evaluated by TF-IDF. We then counted the number of meaningful words with a TF-IDF score of 0, indicating the word has not appeared in the other text, in each type of researcher profile (MeSH-based, abstract-based, and human-written). For this purpose, we used the XML MeSH Dataset [50] collected by the NIH in 2024, ensuring that only words indicative of originality were included.

### 2.3.3 Syntactic Analysis Metrics

We also used lexical and syntactical features to compare the sentence structures within each profile. Specifically, we began with part-of-speech (PoS) tagging and dependency parsing of each sentence in the profiles. Then, we measured the complexity and ambiguity of the sentences in five dimensions: distribution of PoS tags, dependency tree depth, syntactic complexity, syntactic ambiguities, and lexical diversity. The distribution of PoS tags summarizes the frequencies of PoS tags. Dependency tree depth reflects the complexity of sentences, defined as the maximum length of parsing paths in a dependency tree. Syntactic complexity is measured by the average lengths of the parsing paths [51], which also captures the complexity of sentences like dependency tree depth. Syntactic ambiguity refers to the average length of phrases that can be ambiguously parsed as dependencies of different components within the same sentence [52]. Lexical diversity is defined as the number of distinct words associated with the same type of PoS tags [53]. We computed paired t-tests on these metrics, with a p-value of less than 0.05 considered statistically significant.

### 2.3.4 Semantic Similarity Metrics

We also employed BERTScore [54], a semantic similarity metric based on pre-trained contextual embeddings from BERT models, to address the limitations of traditional NLG metrics in capturing true semantic similarity. Traditional metrics rely on exact word matching, while BERTScore calculates similarity as cosine similarity between BERT embeddings for all tokens in the candidate and reference sentences. Hence, BERTScore provides a more powerful and meaningful measurement of semantic similarity. BERTScore precision, recall, and F1 scores were computed for three pairwise comparisons: (1) MeSH-based GPT-generated research summaries versus human-written summaries, (2) abstract-based GPT-generated research summaries versus human-written summaries, and (3) paraphrased research summaries versus human-written summaries. The comparison with paraphrased summaries serves as a validation baseline, because paraphrased summaries should be highly semantically similar to the originals despite low lexical overlap. We used the bert-base-uncased model for all calculations and performed paired t-tests ($\alpha = 0.05$) to assess statistical significance between methods.

## 2.4 Human Evaluation

For human evaluation, we randomly selected 18 researchers and compared LLM-generated profiles with those written by the researchers. The evaluation metrics included overall impression, factual accuracy, granularity of details, readability, comprehensiveness, specificity, and conciseness (Supplementary Tables 1-3). During the evaluation process, participants were presented with three profiles: two generated by LLMs and one by scholars. The order of presentation was randomized to minimize potential order effects. Each dimension was evaluated using a 5-point Likert scale. The evaluation was carried out by four senior team members with experience in writing and reviewing scientific literature. To mitigate individual evaluator bias, each researcher's profile was independently assessed by three evaluators. In addition, we measured inter-rater reliability using Gwet's AC1 coefficient. This chance-corrected measure of agreement is specifically designed to address limitations in kappa statistics [55]. Unlike Cohen's kappa, Gwet's AC1 is usually more robust when rating highly skewed distributions, which were observed in our results.

### 2.5 Recency Sensitivity Analysis

To assess whether recency weighting is necessary, we trained an LDA model (with 30 topics) on all PubMed abstracts. For each researcher, the publication abstract was assigned to its dominant topic for that year, which is defined as the one with the highest posterior topic probability per the LDA topic modeling results. For each year, we identified the set of distinct dominant topics for each author to capture diverse topics that the author published over time. We quantified topic breadth over time using a per-researcher diversity score, calculated by the number of unique topics divided by the number of publications. To make this diversity score metric more robust, we additionally explored three complementary metrics: normalized Shannon entropy, which measures how spread across the publications are in the topics actually used for one author; Hill number, which measures what is the number of topics in which the author's topic mix is as diverse as being perfectly even across; Gini-Simpson index, which measures the probability that two random papers by an author are in different topics. Therefore, for normalized Shannon entropy, higher number indicate the publications are more evenly spread across the topics the author actually uses and smaller number means the topics are concentrated on a small subset of the topics used; while Hill number measures the number of equally used topics that would yield the observed entropy: higher the Hill number is, the broader the researcher has in his/her publications, and vice versa. For the Gini-Simpson index, a higher number indicates a greater topic diversity, and a lower number means the topics are more concentrated. Lastly, higher year-to-year Jaccard similarity numbers indicate a greater persistence of topics from one year to the next, and vice versa.

We then summarized year-to-year shifts using a cohort-level transition matrix. For every pair of adjacent years, we characterized the yearly transition of distribution over dominant topics. To reduce noise and improve readability, we included only authors with at least 10 abstracts and displayed the 15 most frequent topics. The transition matrix is visualized as a single heatmap (Supplementary Figure 1).

To select the number of topics in the LDA model, we evaluated the full pipeline with different K values, ranging from 5, 10, 20, 30, 50, to 100. For each K, we maintained identical preprocessing and author-year aggregation, then evaluated the model fit using log-likelihood and perplexity, and the heatmap coverage by calculating the fraction of all transitions captured by the 15 topics on the heatmap. Three additional metrics introduced for the robustness of the diversity score, as well as the diversity score itself, were also run in K-Sweep for comprehensive analysis.

## 3. Results

We searched self-written research profiles for a total of 595 researchers from Columbia University and downloaded the abstracts of all their PubMed publications. After filtering out those without self-written profiles, we included 167 (28%) of researchers and their profiles that can serve for evaluation purposes.

### 3.1 Comparative Analysis Using Automatic Metrics

Figure 4 shows that MeSH-based or abstracts-based profiles demonstrate low Natural Language Generation (NLG) scores ranging between 0 and 100, with all scores below 15, indicating little vocabulary overlap between machine-generated and human-written summaries. Note that NLG metrics may not precisely reflect semantic similarity or the overall quality of the content, even though they are widely adopted for evaluating word choice similarity. To test this hypothesis, we also calculated the NLG scores for summaries generated by paraphrasing the self-written summaries. Although the paraphrased summaries accurately represent the human-written profiles, the NLG scores are not statistically significantly higher than the other machine-generated summaries. This observation aligns with findings from recent studies [47–49].

We also observe that self-written summaries tend to include newly coined concepts that are unavailable in the summaries generated by machine using scholars' publications. For example, at different times, biomedical scientists have coined concepts such as “learning health systems”, “precision medicine”, and “individualized medicine”. Applying a stop word filter, which removes inconsequential words with little value, we identified 161 distinct concepts used in self-written summaries but absent in machine-generated summaries (Supplementary Table 4). This finding is echoed by the finding of a recent study. To better understand this phenomenon, we assessed the differences in vocabulary usage by computing the KL divergence between their distributions over term frequency. MeSH-based and abstract-based summaries demonstrated KL divergence scores of 8.56 and 8.58, respectively, when comparing their vocabulary distributions against human-written summaries. Recall that important or distinguishing terms are typically assigned higher TF-IDF weights. Machine-generated profiles often contain concepts that differ from those selected by researchers, suggesting the inclusion of potentially irrelevant information or overly specific details. Moreover, the low variance of 0.67 in the KL divergence scores indicates that both MeSH-based and abstract-based summaries consistently deviate from human-written summaries.

### 3.2 Semantic Similarity Analysis

To enhance our lexical analysis, we evaluated semantic similarity using BERTScore (Figure 5). The results show a large contrast with traditional lexical metrics. While BLEU, ROUGE-L, and METOR scores were all below 0.15, BERTScore F1 values were significantly higher. Specifically, MeSH-based GPT-generated summaries had an F1 score of 0.542, abstract-based profiles scored 0.555, and paraphrased summaries achieved 0.851 when compared against human-written summaries. All three comparisons demonstrated statistically significant differences, each with a p-value <0.0005. The moderate BERTScore F1 values (ranging from 0.542 to 0.555) for both machine-generated summaries indicate that these profiles successfully captured semantically related concepts expressed in the self-written summaries, while some topics were still missed when compared to the near-perfect semantic alignment observed with paraphrased summaries. The precision scores exceeded recall scores for both machine-generated summaries methods (MeSH Term-based: 0.584 vs. 0.509; Abstract-based: 0.562 vs. 0.550), suggesting that while the machine-generated content is highly relevant, it lacks specific details present in human-written research summaries. This observation supports our analysis of BERTScores as a meaningful measure of semantic similarity in this context. The high BERTScore observed for paraphrased summaries (F1 = 0.851) validates the metric's ability to capture semantic similarity, even when traditional NLG metrics indicate lexical differences. This finding supports our hypothesis that low lexical scores do not necessarily indicate poor summary quality and highlights the importance of using BERTScore as a complementary evaluation approach.

### 3.3 Syntactic Analysis

To further understand and characterize the differences between machine-generated and human-written profile summaries, we analyzed linguistic and structural patterns, including the maximum depth of dependency trees, syntactic complexity, syntactic ambiguity, part of speech (PoS) distribution, and lexical diversity (Figures 6 and 7). The maximum depth of dependency trees was not statistically different between human-written and machine-generated (MeSH- or abstract-based) summaries. In addition, machine-generated profiles exhibit a similar level of complexity in syntactic patterns as human-written ones. As shown in Figure 6, human-written research summaries have an average syntactic complexity of 3.793. At the same time, machine-generated profiles based on MeSH terms and abstracts exhibit higher average complexity scores of 3.853 and 4.198, respectively. The MeSH-based profiles demonstrate lower syntactic ambiguity, with a score of 4.190, compared to 9.605 for abstract-based profiles and even lower than 5.720 for the self-written summaries. The top panel of Figure 7 shows that human-written and machine-generated profiles have similar patterns of PoS distributions, where nouns are the most frequently used type of words, followed by adjectives, appositions, and verbs. As shown in the bottom panel of Figure 7, MeSH-based profiles are less lexically diverse. In contrast, abstract-based profiles have a similar lexical diversity as compared to the human-written ones.

### 3.4 Human Evaluation

Figure 8 shows the evaluation results for the distribution of the factual accuracy, granularity, conciseness, readability, comprehensiveness, specificity, and overall impression. The complete survey results are shown in Supplementary Tables 1-3. The overall Gwet's AC1 coefficient [55] is 0.634. We note that most disagreements between evaluators occurred in summaries rated as low quality, ranging from fair to very poor. Despite a lack of agreement among evaluators, their ratings consistently reflected negative sentiment. We showed that inter-annotator reliability was higher for summaries of better quality. For example, Gwet's AC1 coefficient for summaries with at least a good overall impression was 0.762, as shown in Figure 9.

### 3.5 Topic Stability Over Time

Across 167 researchers, publication topics remain relatively consistent for the majority. Specifically, 80 researchers (48.8%) had diversity scores below 0.3, indicating focused research interests, while only 7 researchers (4.3%) exhibited substantial topic evolution, with diversity scores above 0.7. The heatmap (Supplementary Figure 1) further confirms that researchers generally remained focused on the same topic over years.

To further verify our observation that researchers tend to stay focused on certain areas throughout their career and that recency weighting has a moderate impact on profiling, we analyzed topic persistence and breadth for each author across a range of LDA topic granularities (K = {5, 10, 20, 30, 50, 100). The mean Jaccard value, indicating year-to-year topic overlap, was high across different topic numbers. As the topic number decreased, although overlap declined, it remained well above zero, indicating substantial continued topic focus, especially at the level of broad areas (Supplementary Table 5).

With K=30, the mean normalized Shannon entropy was 0.786, the mean Hill number was 5.59, and the mean Gini-Simpson index was 0.692, indicating researchers typically work within a few topics rather than spreading uniformly across a wider range of topics (Supplementary Table 5).

### 3.6 Topic Number (K) Sensitivity Analysis

Across the k value of 5, 10, 20, 30, 50, 100, perplexity improved from 618 (K=5) to 548.6 (K=30) and then worsened as the value of K increases (Supplementary Figure 2), with log-likelihood following a similar pattern where it is the best when K=30 (Supplementary Figure 3). Importantly, heatmap coverage by the top 15 topics falls drastically beyond K=30 (K=20: 0.80; K=30: 0.44; K=50: 0.21; K=100: 0.11), indicating that higher K produces very sparse, less interpretable transition views (Supplementary Figure 4). Based on the above analysis, we chose K=30 topics for the LDA model.

## 4. Discussion

This study leverages LLMs to summarize researchers' interests and generate narrative researcher profiles based on their PubMed publications. We systematically compared the resultant summaries to the profiles written by the researchers. Based on the comparison, we identified lexical and semantic differences but similar language styles between machine-generated and human-written profiles.

First, we identified the varying word choices between machine-generated and human summaries, which were reflected in the low BLEU, ROUGE-L, and METEOR scores. We acknowledge that although these NLG metrics have been widely adopted for assessing the quality of machine-generated content, such metrics overly rely on common word sequences or stems and do not comprehensively reflect the text quality. To confirm this, we compared the human-written profiles against the paraphrased version. The paraphrased profiles were semantically close to the human-written profiles but still demonstrated low NLG metric scores. The limitations of the NLG metrics highlight the imperative need for inventing more robust evaluation metrics in the future.

To address such limitations in traditional NLG metrics, we used BERTScore, a metric for embedding-based semantic similarity evaluation. BERTScore analysis provided important insights: although lexical overlap between machine-generated and human-written summaries was low, intermediate F1 scores (0.542-0.555) indicated that machine-generated summaries were able to capture related concepts, even when phrasing differently. This gap between semantic and lexical similarity supports our finding that, although machine-generated summaries can identify crucial aspects and concepts, they tend to stay closer to the source vocabulary and lack the conceptual abstraction found in human writing. The validation using paraphrased summaries (BERTScore f1 = 0.851) further confirmed that high semantic similarity can exist alongside substantial lexical variation, showing the limitation of traditional NLG metrics and the necessity of adopting complementary evaluation metrics. The nearly 30% gap between the BERTScore between both machine-generated summaries and paraphrased summaries demonstrates that, while current LLMs can capture related concepts, they still struggle with the level of abstract synthesis characteristic of human authors. This finding presents a key challenge in automated research profiling and suggests that future improvements should focus on closing the gap between current LLM capabilities and human-like abstraction.

In addition to the widely used NLG metrics, a broader concern is homogenization, or lack of novelty in LLM generated texts [34, 35]. For scholar profiling, this risk argues for novelty-aware evaluation, so distinct contributions are not washed out. Therefore, we systematically captured the semantic differences between the human-written profiles and LLM-generated summaries by comparing the vocabulary distribution characterized using TF-IDF distribution, where key terms are typically assigned high weights. Using the KL divergence of vocabulary distribution, we identified a divergent preference for keywords, which signify the core topics of researchers' interest (Figure 10). As such, we inferred that the lack of overlapping terms is not confined to trivial words. We identified a total of 161 distinct terms that only appear in human-written summaries. Even though we derive researcher profiles directly from publication abstracts or MeSH terms, human-generated summaries contain many exclusive MeSH terms not represented in machine-generated summaries, highlighting that human writers are more adept at abstracting and summarizing nuanced text. At the same time, LLMs tend to repeat input at the expense of more nuanced or personalized language. For example, in a researcher's profile (Figure 3), nuanced descriptions such as 'develop neuro-symbolic methods to automate medical evidence comprehension (making PubMed computable)' illustrate an advanced synthesis of methodologies and goals. In contrast, the abstract-based LLM-generated summary lists granular methodologies such as 'natural language processing (NLP),' 'evidence retrieval,' and 'artificial intelligence (AI)' without synthesizing these into integrated concepts or emphasizing their application context clearly. This tendency towards verbatim repetition rather than abstraction illustrates the limitations of current LLM-generated profiles. This observation is also reflected in the higher lexical diversity scores of the human-written profiles, where human authors frequently weave interpretive or subjective descriptions—an element of originality that the model does not emulate well. These patterns also reflect findings that AI-assisted essays converge on common wording and topics, producing within-group homogeneity [34, 35].

Our manual evaluation studies based on expert survey results further confirm the observations from the automated evaluation of lexical and semantic differences. Besides the higher rating of overall impression, the human summaries were consistently rated higher in all aspects of summary quality, including comprehensiveness, factual accuracy, and others. Notably, human summaries dominate both comprehensiveness and conciseness, indicating that the LLM approach of stitching details scattered in input sources does not guarantee full coverage of key information and may include excessive details such as ‘through models like PICOX for extracting PICO entities and normalizing complex interventions.’ This also confirms our observation that human-written and machine-generated summaries emphasize different keywords, where keywords in the human summaries could be crafted or abstracted instead of copied from the input.

Despite the above-mentioned differences, machine-generated and human-written summaries demonstrate similar language patterns. They contain sentences with a maximum dependency tree depth of 8.6 and present similar syntactic complexity. Furthermore, they exhibit a similar distribution across various types of PoS categories. The only exception is the syntactic ambiguity, where human-written and MeSH-based profiles are lower than abstract-based ones. This is a side effect of LLM behavior, where they commonly verbatim repeat phrases in the text summarization, concatenating scattered information, which can produce ambiguous expressions.

Having established overall stability and robustness, we next examine how different textual inputs affect profile quality. MeSH-based profiles are rated slightly higher than abstract-based ones. Note that abstracts contain more detailed information than MeSH terms. However, the large volume of publications and the limit of the LLM context window, i.e., the maximum number of tokens that LLMs can process for one request, pose a challenge to directly using the full-text publications as input. Using MeSH terms for profile generation circumvents the LLM context window limitation and demonstrates competent performance compared to the summarization approach using abstracts. This highlights keyword-based text generation as a promising approach for profiling scholars' research interests. Consistent with our topic-evolution analysis across 167 researchers, most researchers maintained stable interests over the past decade (Supplementary Figure 1); accordingly, we weighted publications equally across years when generating profiles, while it is worth noting that recency-weighted variants may benefit the small subset with marked topic shifts.

Finally, to make sure that the diversity ratio (unique topics/publications) defined in this paper does not underestimate diversity for highly productive authors with a wide range of subjects of research, as the numerator is controlled by K while the denominator increases with publication count, we strengthened the measurement of diversity through three length-robust measures: normalized Shannon entropy, the Hill Number, and the Gini-Simpson index. At K=30, means of cohort means were 0.786 (Shannon), 5.59 (Hill), and 0.692 (Gini-Simpson). These results indicate that authors tend to concentrate their activities within approximately 5-6 effective topics on average, which is consistent with the concept of concentrating on persistent topics rather than switching between wide-ranging topics. Across K from 5 to 100, these measures varied smoothly without contradicting the conclusion that we drew. Year-to-year Jaccard results have consistently high numbers over time (Supplementary Table 5). Collectively, these analyses eliminate the length-bias constraint and substantiate that most researchers spend most of their time in a small number of topics and switch fields far less often, making the recency of publication less impactful to profiling.

This study has several limitations. First, the publication record collection process uses heuristics to determine the relevance and significance of the author's contributions based on the authorship orders (e.g., the first three authors and senior). This step can be further improved to become more systematic and automated. Second, we used institutional affiliation to disambiguate publications from different scholars with the same name. This could be further enhanced by integrating an external knowledge base of scholar affiliation and expertise or a previously published, more sophisticated algorithm for researchers' name disambiguation [56]. Third, we may have inadvertently introduced potential selection bias by restricting the dataset to the last 10 years and publications in the first three or senior authorship positions for each researcher. This filter could exclude influential older publications or significant middle-author contributions—particularly in fields or big projects where collaboration or multi-authorship is common. This potential bias could undermine the comprehensiveness and representativeness of generated research summaries. Fourth, the data sources for generating research summaries were restricted to our institutional college of physicians and surgeons, chosen for our familiarity with them to facilitate human evaluation. This cohort may not fully represent the comprehensive research topics across other disciplines or institutions. As a future direction, we can extend the study to include more diverse disciplines and institutions to evaluate the two LLM-based approaches to profiling scholars. Finally, our study has a relatively small sample size of 18 researchers in the human evaluation phase. Although expanding the human evaluation to a larger set of researcher profiles would undoubtedly improve the robustness and generalizability of our findings, practical constraints such as the long time required to distribute surveys, collect responses, and analyze data prevented us from doing so within the available timeframe. Future research should aim to conduct human evaluation on larger samples to confirm the findings in this study.

## 5. Conclusions

This study discusses the capabilities and limitations of using LLMs to summarize scholars' research interests. We explore two approaches, i.e., text summarization using publication abstracts and text generation using MeSH terms from publications. We conducted a systematic evaluation using widely adopted NLG metrics, lexical and syntactic patterns, and expert surveys. Our results show that machine-generated summaries emphasize different keywords than human-written summaries, which still leaves room for further improvement in the research interest profiling. Despite the limitations, our study demonstrates the potential of LLMs to facilitate scholar profiling. Directions of future work include fully automating publication screening and name disambiguation for researchers from different institutions and backgrounds but with the same names, using retrieval-augmented language models with external knowledge bases.

**Funding Sources**

This work was supported by the National Center for Advancing Translational Sciences (NCATS) of the National Institutes of Health (NIH) under grant number UL1TR002384. This research was funded by National Institute of Health grants R01LM014344 and R01LM014573, and National Library of Medicine grant T15LM007079.

## Data availability

The data underlying this article will be available upon request.

## Code availability

The code will be available upon request.

## Contributorship

YL: data curation, formal analysis, investigation, methodology, validation, visualization, writing - original draft;

GZ: conceptualization, data curation, formal analysis, investigation, methodology, validation, project administration, writing - original draft;

ES: data curation, formal analysis, investigation, methodology, validation, writing - review & editing;

YF: conceptualization, data curation, formal analysis, investigation, methodology, validation, writing - review & editing;

FC: conceptualization, data curation, formal analysis, investigation, methodology, validation, writing - review & editing;

BI: investigation, validation, writing - review & editing;

YP: formal analysis, investigation, methodology, validation, visualization, supervision, funding acquisition, writing - review & editing;

CW: conceptualization, data curation, formal analysis, investigation, methodology, validation, project administration, supervision, resources, funding acquisition, writing - review & editing.

**Figure 1:** The overview of our proposed method to generate the researcher profiles.

**Figure 2:** A Prompt example for research profiling.

**Figure 3:** Examples of MeSH-based, abstract-based, and paraphrased LLM-generated researcher profiles and the human-written profiles.

**Figure 4:** Comparison of machine-generated research profiles using MeSH Terms, abstracts, and human-written profiles using Natural Language Generation metrics. Significance Legend: ns: $p>=0.05$; *: $0.01 \le p < 0.05$; **: $0.001 \le p < 0.01$; ***: $p < 0.001$

**Figure 5:** BERTScore evaluation results comparing machine-generated profiles with human-written profiles. (a) Bar chart showing precision, recall, and F1 scores for MeSH-based, abstract-based, and paraphrased summaries. (b) Box plot showing F1 score distributions across 167 researchers.

**Figure 6:** Comparison of machine-generated research profiles using MeSH Terms, abstracts, and human-written profiles using syntactic analysis. Significance Legend: ns: $p>=0.05$; *: $0.01 \le p < 0.05$; **: $0.001 \le p < 0.01$; ***: $p < 0.001$

**Figure 7:** Frequency percentage of PoS tag as a measure of PoS distribution (a) and Lexical Diversity (b). Noun (NOUN), Adjective (ADJ), Adverb (ADV), Verb (VERB), Auxiliary Verb (AUX), Pronoun (PRON), Adposition (ADP), Punctuation (PUNCT), Determiner (DET), Coordinating Conjunction (CCONJ), Subordinating Conjunction (SCONJ), Particle (PART), Interjection (INTJ), space (SPAåCE), Numeral (NUM), Symbol (SYM), Proper Noun (PROPN), and Other (X).

**Figure 8:** Survey results for Factual Accuracy (a), Granularity (b), Conciseness (c), Readability (d), Comprehensiveness (e), Specificity (f), and Overall Impression (g) for human-written researcher profiles, MeSH Term-based GPT-generated researcher profiles, and abstract-based GPT-generated research summaries.

**Figure 9:** Gwet AC1 score for low, middle, and high overall impression by evaluators in surveys for evaluation of human-written researcher profiles, MeSH Term-based GPT-generated researcher profiles, and abstract-based GPT-generated research summaries.

**Figure 10:** An example human-written profile, abstract-based machine-generated profile, and MeSH Term-based machine-generated profile of a researcher with unique keywords in the human-written profile highlighted in red and keywords from publication records highlighted in blue.